\documentclass[letterpaper, 10 pt, conference]{ieeeconf}  

\IEEEoverridecommandlockouts                              

\usepackage{graphicx}
\usepackage{booktabs}
\usepackage{url}

\title{\LARGE \bf
Towards Considerate Human-Robot Coexistence: A Dual-Space Framework of Robot Design and Human Perception in Healthcare
}

\author{Yuanchen Bai$^{1,2}$, Zijian Ding$^{3}$, Ruixiang Han$^{1,2}$, Niti Parikh$^{1}$, Wendy Ju$^{1}$ and Angelique Taylor$^{1,2}$
\thanks{*This material was supported by the National Science Foundation under Grant No. IIS-2423127.}
\thanks{$^{1}$Yuanchen Bai, Ruixiang Han, Niti Parikh, Wendy Ju, and Angelique Taylor are with Cornell Tech, New York, USA
        {\tt\small yb299@cornell.edu, }{\tt\small rh652@cornell.edu, }{\tt\small ntp27@cornell.edu, }{\tt\small wendyju@cornell.edu, }{\tt\small amt298@cornell.edu}}%
\thanks{$^{2}$Yuanchen Bai, Ruixiang Han, and Angelique Taylor are also with the Department of Information Science, Cornell University, Ithaca, USA}        
\thanks{$^{3}$Zijian Ding is with University of Maryland, College Park
        {\tt\small ding@umd.edu}}%
}

\begin{document}

\maketitle
\thispagestyle{empty}
\pagestyle{empty}

\begin{abstract}
The rapid advancement of robotics is reshaping what it means for humans and robots to coexist---through expanded capabilities, more intuitive interactions, and deeper integration into real-world workflows. Beyond sharing physical space, this coexistence is increasingly characterized by organizational embeddedness, temporal evolution, social situatedness, and open-ended uncertainty. 
Because such coexistence extends beyond a single encounter, understanding healthcare robots requires looking beyond initial acceptance to how stakeholders' perceptions evolve through continued engagement. Yet, prior work has largely relied on single-point snapshots of attitudes and acceptance, offering limited insight into coexistence as a long-term, dynamic process.
We address these gaps through in-depth follow-up interviews with nine participants from a 14-week co-design study on healthcare robots. We identify the \textbf{human perception space}, which includes four interpretive dimensions (i.e., degree of decomposition, source of evidence, scope of reasoning, and temporal orientation). We enrich the conceptual framework of human-robot coexistence by conceptualizing the mutual relationship between the human perception space and the robot design space as a \textbf{co-evolving loop}, in which human needs, design decisions, situated interpretations, and social mediation continuously reshape one another over time.  Building on this, we propose \textbf{considerate human-robot coexistence}, arguing that humans act not only as design contributors but also as interpreters and mediators who actively shape how robots are understood and integrated across deployment stages. Our related prior work and supplementary materials, including the interview protocol, are available at \url{https://byc-sophie.github.io/considerate-human-robot-coexistence/}
\end{abstract}

\section{Introduction \& Related Work}


The advancement of robotics is reshaping how humans and robots coexist, manifesting across three interrelated dimensions. At the capability level, advances in large language models and agentic Artificial Intelligence (AI) are expanding robots' reasoning, planning, and multimodal perception capabilities~\cite{durante2024agent}. At the interaction level, these expanded capabilities enable more intuitive language-based interaction~\cite{lu2023extracting} and more expressive robot behaviors (e.g., social nodding and multimodal feedback)~\cite{mahadevan2024generative}. At the contextual level, robots are increasingly envisioned in real-world workflows, moving from backend support into frontline participation and from standalone tools to collaborative team members~\cite{bai2026towards}.

Against this backdrop, \textit{coexistence} has emerged as a crucial lens for understanding human-robot relationships. However, the term carries distinct connotations across disciplines and often varies by context, lacking a unified definition \cite{knox2021usage}. 
In ecology and the social sciences, for instance, coexistence describes the sustained co-presence of different actors sharing resources and environments, which is an inherently ongoing, dynamic process rather than a fixed arrangement \cite{armstrong1976coexistence,loring2016toward}.
Within existing human-robot interaction (HRI) and robotics research, coexistence has been understood more as a state of humans and robots sharing physical space, with a primary focus on safety, efficiency \cite{magrini2020human, merckaert2022real}, and spatial organization, such as human-aware social navigation \cite{zhang2025human} or layout design in shared environments \cite{zhi2021designing}. 
Yet, this spatially-centered conception is increasingly inadequate in light of the three-level shift aforementioned.
In our prior co-design study,
participants' understandings of robots evolved as they moved from
abstract ideation to high-fidelity prototyping over a 14-week process
~\cite{bai2026towards}.
Human-robot coexistence is therefore more than a spatial state; it is an evolving process shaped by multiple actors over time.

Here, we enrich the conceptual understanding of \textbf{human-robot coexistence} through the following four features:
(1)~\textit{organizational embeddedness}: robots act as participants integrated within real-world workflows and organizational structures (e.g., crash cart robots in resuscitation teams \cite{tanjim2025human});
(2)~\textit{temporal evolution}: human-robot relationships evolve over time as users accumulate experience and continuously update their understanding of robot capabilities \cite{bai2026towards};
(3)~\textit{social situatedness}: the roles and meanings of robots are shaped not only by their technical capabilities, but also by stakeholders' interpretations and the social contexts in which they are deployed \cite{mahadevan2024generative};
(4)~\textit{open-ended uncertainty}: robot interaction unfolds in dynamic and uncertain environments, requiring adaptive reasoning that transcends pre-defined scripts, particularly in complex domains such as healthcare \cite{taylor2020situating}.

Under this richer conception, human-robot coexistence is not merely a matter of whether a robot is accepted at a single point in time, but also of how stakeholders understand, interpret, and re-evaluate robots as coexistence unfolds over time. Human perception therefore becomes a crucial variable.
Yet, existing work has largely focused on static assessments of attitudes or acceptance towards healthcare robots. Such studies often explore how people anticipate teaming up with robots in imagined scenarios \cite{abrams2025teaming} or rely on qualitative interviews to catalog perceived potentials (e.g., reducing burden) and concerns (e.g., technical unreliability) \cite{smola2025attitudes}. 
While these works reveal what stakeholders think about robots, they offer limited insight into how these perceptions form, interact, and evolve over time.
A set of questions therefore remains underexplored: how stakeholders' impressions of robots form and through what attributions, how these perceptions evolve as exposure deepens, and what active roles humans play in shaping coexistence beyond passively evaluating it.

To address these gaps, we build on our prior co-design work \cite{bai2026towards} and pivot from robot design to human perception. We conducted follow-up interviews with nine participants who had completed a 14-week co-design study (from abstract ideation to high-fidelity prototyping), examining how their assessments of the long-term promise of healthcare robots evolved through sustained engagement and how they envisioned human-robot coexistence in care settings.

This paper addresses the following research questions:

\begin{itemize}
\item \textbf{RQ1:} How do stakeholders' evaluations of a healthcare robot's promise evolve throughout a long-term co-design process, and what underlying factors drive these attitudinal shifts?
\item \textbf{RQ2:} How do participants conceptualize future human-robot coexistence, what idealized states of coexistence do they envision, and what roles do humans assume within this paradigm?
\end{itemize}

\begin{figure}[t]
  \centering
  \includegraphics[width=0.45\textwidth]{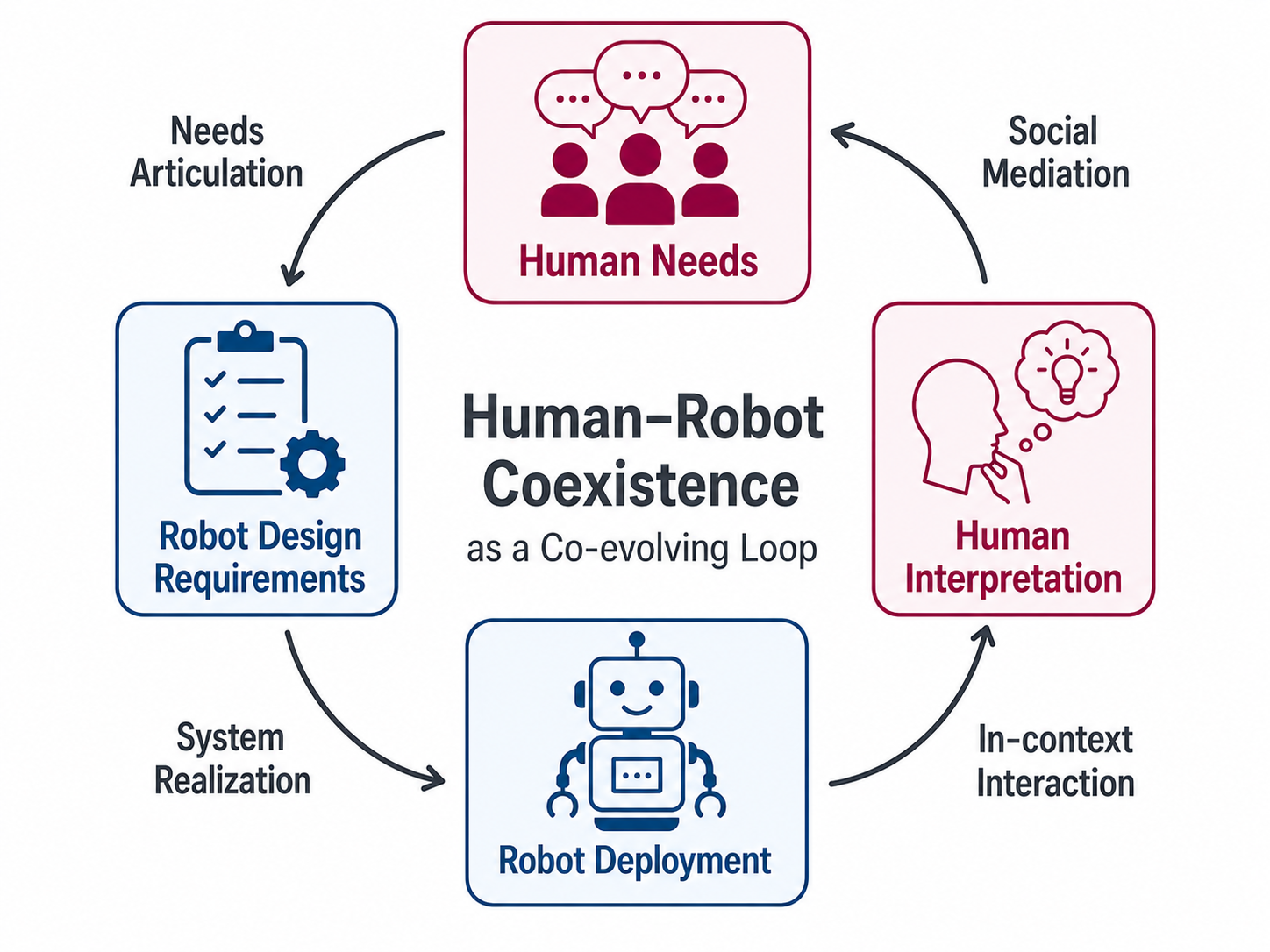}
    \caption{
We conceptualize human-robot coexistence as a \textbf{co-evolving loop} between robot design and human perception.
\textbf{Human needs} are articulated into \textbf{design requirements}, which are realized into \textbf{robotic systems deployed in context}. 
Through deployment and situated interaction, these systems shape \textbf{human interpretation}, which is further propagated via social mediation, feeding back into evolving human needs. 
This loop highlights the dynamic and reciprocal process through which human-robot coexistence is continuously formed and reshaped over time.
}
  \label{fig:loop}
\end{figure}

The central contribution of this paper is to extend existing understandings of human-robot coexistence by conceptualizing it as a co-evolving loop between robot design and human perception.
Concretely, this paper makes three contributions:
\begin{itemize}
    \item \textbf{Human perception space.}
    We identify the human perception space---the previously underexamined,
    human-side counterpart to the \textbf{robot design space} established in
    our prior work \cite{bai2026towards}---through four interpretive dimensions:
    \textit{degree of decomposition}, \textit{source of evidence}, \textit{scope of reasoning} and \textit{temporal orientation}. These dimensions form a multi-faceted continuum
    that shapes how stakeholders organize information, attribute risks, and
    evaluate the functional and normative boundaries of healthcare robots in
    practice (\textit{see Fig.~\ref{fig:spaces}}).

    \item \textbf{Dual-space co-evolving loop.}
    We further show that these two spaces are not static counterparts, but mutually shape one another through a \textit{co-evolving loop},
    in which human needs are articulated into design requirements, realized
    as deployed systems, and continuously reshaped through in-context
    interaction, interpretation, and social mediation
    (\textit{see Fig.~\ref{fig:loop} and Fig.~\ref{fig:spaces}}).

    \item \textbf{Design implications for long-term coexistence.}
    Using \textit{considerate} coexistence as a lens, we derive implications
    for its realization across deployment stages, arguing that humans act not
    only as \textit{design contributors} who articulate needs and shape
    robotic systems, but also as \textit{interpreters and mediators} who
    continuously influence how robots are understood and integrated in
    real-world environments. We highlight making design rationales legible
    before deployment, supporting ongoing interaction and mediation during
    use, enabling gradual engagement, and respecting professional boundaries.
\end{itemize}

\section{Methodology}
To examine whether and how participants' perceptions of healthcare robots have evolved following an extended co-design process, and how they envision human-robot coexistence in care settings, we adopt a qualitative approach based on in-depth semi-structured interviews, incorporating a retrospective attitudinal assessment. The study was IRB approved (IRB \#: IRB0145631).

\textbf{Study Context.} Participants had previously completed a 14-week co-design process across
three healthcare settings: an emergency department, a sleep disorder clinic, and a long-term rehabilitation facility
\cite{bai2026towards}. The process progressed from needs identification and iterative prototyping to high-fidelity, full-scale robot prototypes envisioned as active participants in care workflows. Participants also attended educational sessions on healthcare workflows, the history of
robotics, and relevant robot technologies. These sessions fostered a shared understanding among stakeholders with different backgrounds and supported an informed co-design process.

\textbf{Participants.} We recruited nine participants (5 male, 4 female; details in Table~\ref{tab:participants}) who had completed the co-design process described above. They spanned diverse backgrounds, including healthcare
workers (HCWs), patients, engineers, makers, and artists, offering heterogeneous perspectives on healthcare robots. Participation was voluntary and uncompensated.

\begin{table}[t]
\caption{Participant demographics and background.}
\label{tab:participants}
\centering
\begin{tabular}{llll}
\toprule
\textbf{ID} & \textbf{Gender} & \textbf{Age} & \textbf{Background} \\
\midrule
P1 & Male   & 30--40 & Programmer/Maker \\
P2 & Male   & 60--70 & Maker/Patient \\
P3 & Male   & 60--70 & Artist \\
P4 & Female & 50--60 & HCW/Artist \\
P5 & Male   & 20--30 & Programmer/Maker \\
P6 & Female & 60--70 & Artist \\
P7 & Male   & 20--30 & Programmer/Maker \\
P8 & Female & 30--40 & Maker \\
P9 & Female & 30--40 & HCW/Programmer \\
\bottomrule
\end{tabular}
\end{table}

\textbf{Interviews.} To capture perception shifts, we asked participants to retrospectively rate their outlook on healthcare robots before and after the co-design experience on a 5-point scale (1 = not promising, 5 = very promising). 
Given the subjectivity of retrospective ratings, we focus on \textit{directional trends} (increase, decrease, or no change) rather than absolute score differences. 
Each interview lasted approximately 30 minutes and was structured around two guiding topics: \textit{Evolution of Perceived Promise}, which investigates the factors
that lead participants to view healthcare robots as more (or less) promising and the underlying reasons for these attitudinal shifts, and \textit{Envisioned Human-Robot Coexistence}, which explored expectations and actionable strategies for coexistence in care settings.

\textbf{Data Collection and Analysis.} The first author conducted all interviews, recorded audio, and transcribed and verified all recordings. 
Transcripts were analyzed using thematic analysis~\cite{clarke2017thematic}. 
The first author developed the initial codes; the third author independently coded a subset of the data. 
Following collaborative discussions, a shared codebook was developed, and both authors applied it to the full dataset. 
We computed Cohen's Kappa to assess the consistency: for RQ1, three themes with five sub-themes ($\kappa = 0.82$); for RQ2, three themes with six sub-themes ($\kappa = 0.86$), indicating high coding consistency.

\section{Results}

\subsection{RQ1: Perceptual Shifts in Healthcare Robots' Promise: Increase, Stability, and Calibrated Decrease}
We observed changes in participants' perceptions of the promise of healthcare robots: five increased, three remained stable, and one decreased.

\textbf{Theme 1: Increased Perceived Promise Through Contextual Validation and Demystification.}  
 Among participants in this group, two recurrent patterns emerged.

\textbf{a) Contextual validation of real needs.}
Interactions with HCWs shifted three participants' views by revealing concrete needs rather than abstract assumptions. For example, P7 found robots more promising after observing moments when HCWs ``could have benefited from robotic assistance''; P9 similarly shifted after recognizing concrete ``potential uses'' in care work.

\textbf{b) From uncertainty to feasibility awareness.}  
Two participants moved from vague uncertainty to a more grounded understanding of what robots could realistically achieve. Through technical demonstrations, P9 came to understand how large language models and machine learning algorithms could enhance robot capabilities and enable more seamless movements. Similarly, P5 reflected: ``[Previously,] I didn't know people were developing robots for healthcare settings. I felt it was doable, but I didn't know if people were actually working on it.'' Through the workshop, P5 gained clarity about ongoing development in this area, which made healthcare robots appear more feasible.

\textbf{Theme 2: Stable Perception Anchored by Pre-existing Frameworks.}  
Stable perceptions appeared to be supported by participants' prior experience and structured knowledge.
\textbf{a) Experience-based anchoring.}  
Three participants anchored their perceptions in prior experience. P1, with a strong technical background and direct experience across multiple healthcare facilities, could readily identify situations where robots would be useful, such as ``filling out forms'' and ``reducing errors or miscommunication.'' He also noted that surgical robots can support precision by helping ``correct how a doctor moves their hands.'' P2, a patient at a long-term rehabilitation facility, extended lived experience with an assistive wheelchair to healthcare robots, viewing similarly embodied systems as everyday support.

\textbf{b) Knowledge-based reasoning.}  
Two participants drew on structured knowledge to frame challenges as solvable rather than discouraging. P1 framed technological maturity as an ongoing process, suggesting that systems can progressively improve over time: ``Turning immature technology into mature technology is exactly what you're doing. It doesn't just happen.'' He thus viewed current limitations as part of a trajectory toward greater capability. Similarly, P6 contextualized safety concerns by comparing robot errors to human fallibility: ``People can also pick up the wrong thing and give it to somebody. Some employees are not trustworthy.'' Rather than viewing mistakes as a reason for concern, she emphasized that systems ``just need proper protocols to function properly.''

\textbf{Theme 3: Calibrated Decrease as Grounded Reassessment.}  
One participant showed a slight decrease in perceived promise, reflecting a shift from idealized expectations to a more realistic appraisal of constraints such as cost. P4 explained: ``It got reduced because I came in imagining robots doing everything. My imagination was outside the frame of reality.'' She added, ``I bring it down to a four because there are real possibilities, but you have to put reality into it.'' Rather than indicating disillusionment, this recalibration may reflect a more grounded engagement.

\subsection{RQ2: Human-Robot Coexistence in Healthcare}

Rather than evaluating robots as isolated technologies or expressing
simply positive or negative attitudes, participants interpreted them
through familiar technological trajectories, institutional signals, and
situated expectations about care. These accounts frame coexistence as a
contextual, evolving process that requires careful deployment.

\textbf{Theme 1: Normalization through Technological Trajectories.}
Five participants framed robots as part of a broader pattern of
technological adoption, suggesting that unfamiliarity may diminish with
repeated exposure. P1 compared robots to early mobile phones: ``it's like the first time, people just bought it to show off, but now it's weird not to have a smartphone.'' P3 similarly emphasized gradual adaptation through everyday infrastructures, noting subway systems where people now ``get used to these things.'' P9 framed acceptance as a matter of time and repeated interaction: ``interacting with them more makes people feel more comfortable.'' Together, these accounts framed coexistence as an extension of ongoing sociotechnical adaptation.

\textbf{Theme 2: Robots as Institutional Signals.}
Seven participants described how robots could shape perceptions of hospitals regarding quality, advancement, and safety, even among those not directly interacting with them. P1 noted that seeing a robot would suggest ``cutting edge technology'', and P3 associated it with ``a better place'' and ``a high-level care system.'' P7 linked robot presence to safety: ``the hospital is advanced enough'' and that robots might ``check things more than once.'' P5 anticipated comparative judgments, suggesting that in the future, hospitals without robots might be perceived as lower quality. At the same time, participants acknowledged subjectivity: as P6 noted, perceptions ``really depend on your bias as a person.''

\textbf{Theme 3: Strategies for Supporting Human-Robot Coexistence.}
Participants articulated practical strategies for supporting coexistence in real-world care settings.

\textbf{a) Social blending.}
Three participants emphasized that robots should blend into care environments socially and aesthetically. P3 highlighted environmental fit: ``in a sleep environment, you want it to be calming. Natural materials would be nicer for people.'' P6 suggested making robots feel ``less like plastic and more organic to give a comforting feeling''. Institutional alignment also emerged as important, with P9 proposing ``giving the robot a uniform.''

\textbf{b) Respecting user preferences.}  
Four participants emphasized that robots should not be uniformly imposed on all patients. Instead, they highlighted accommodating individual comfort levels through pre-screening and flexible alternatives. As P5 suggested, patients could be asked, ``are you comfortable with robots---yes or no? If not, they would not be included.'' Similarly, P9 noted that ``we can offer an option if some people need more time to adapt to the robots.''

\textbf{c) Human mediation as a bridge.}
Five participants highlighted the role of humans in introducing and contextualizing robots. P1 described incorporating robots into communication practices: ``tell a patient that we have this robot here. It does all these things and everything is secure.'' P4 emphasized clarifying intent, for example, explaining that robots are ``not there to replace you, but to assist you''. Participants also stressed education and staged introduction: P8 noted that ``you need to educate the public. Whenever something new is introduced, there's always going to be misconceptions about it,'' while P7 suggested ``a test run to see how people are feeling.''

\textbf{d) Maintaining professional and contextual boundaries.}  
Seven participants emphasized that robots should not be introduced for their own sake, but to meaningfully improve care and workflows. As P5 noted, ``the point of a hospital is to take care of patients, not to teach them that robots are cool.'' Accordingly, they held that robots should operate within clearly defined professional boundaries, justified by their functional contribution. Participants also stressed the continued importance of human oversight in complex care contexts. As P4 illustrated, ``critical conditions'' can be overlooked without human judgment, pointing to perceived limits of purely automated systems.

\section{Discussion} 

\begin{figure*}[t]
  \centering
  \includegraphics[width=0.71\textwidth]{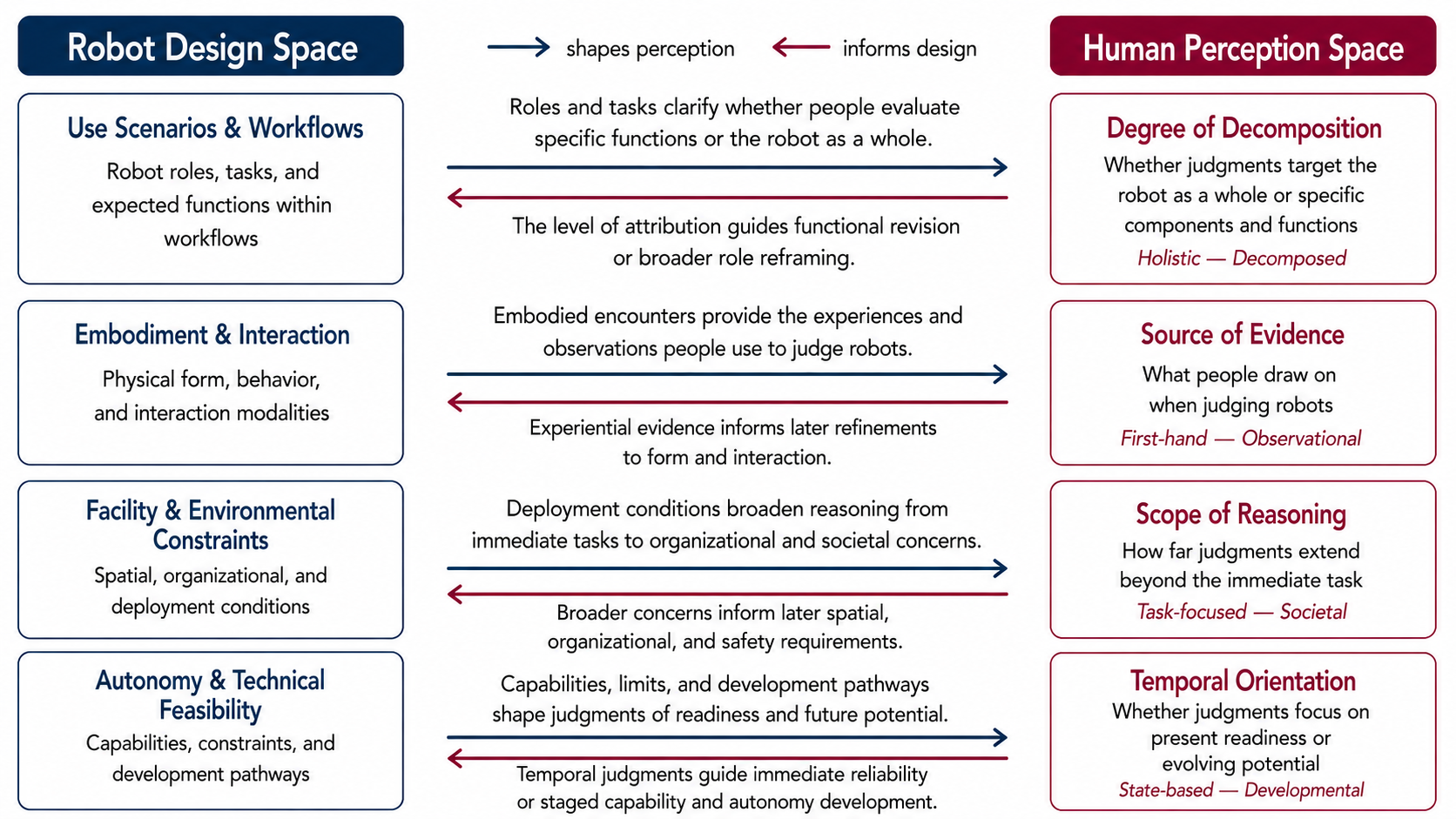}
\caption{\textbf{Robot design and human perception spaces for human-robot
coexistence.} The left side summarizes key design considerations, and
the right side characterizes four interpretive dimensions of human
perception. Blue arrows indicate how design considerations shape
perception, while red arrows indicate how human interpretations inform
later design. The aligned pairings highlight recurring cross-space
relations rather than exhaustive or exclusive one-to-one mappings.}
  \label{fig:spaces}
\end{figure*}

\subsection{Human Perception Space: Four Interpretive Dimensions}

We found that participants interpreted robots through a set of \textbf{interpretive dimensions} that shaped how they organized information, attributed risks, and judged what robots could or should do in care settings. These dimensions suggest a continuous perception space (see Fig. \ref{fig:spaces}).

\textbf{(1) Degree of Decomposition.}
This dimension captures whether participants interpret the robot as a \textit{decomposable system} or a \textit{holistic} technological entity.
At the decomposed end, participants distinguished among sources of problems, including hardware, network, and data infrastructure. P1 illustrated this distinction by attributing privacy concerns to broader infrastructure rather than to ``the robot'' as a whole. At the holistic end, one unreliable aspect could generalize to the robot as a whole: P2's concerns about data accuracy reflected worry about the robot as a concept rather than about specific sources, such as incorrect data input or analysis algorithms. This dimension thus concerns whether problems are framed as localized, debuggable issues or as global uncertainty about ``the robot'' itself.

\textbf{(2) Source of Evidence.}
This dimension captures the evidence participants drew upon, ranging
from \textit{first-hand experience} to \textit{observational or socially
derived} evidence. Regarding the former, P2 drew on lived experience
with an assistive wheelchair, and P1 grounded judgments in direct
exposure to healthcare technologies. At the other end, P9
reflected that seeing others grow excited about robots shaped their own
sense of the technology's promise, while acknowledging the possible bias
during observation. This distinction echoes prior work showing that media and vicarious exposure can shape robot perceptions even without direct contact~\cite{sundar2016hollywood}.

\textbf{(3) Scope of Reasoning.}
This dimension describes how broadly participants framed their evaluation, from \textit{task-focused} to \textit{societal-level} reasoning.
At the task-focused end, participants evaluated specific functions such as patient
guidance and supply delivery. At the broader end, participants situated robots within wider concerns such as professional roles, organizational incentives, and job displacement. For example, P3 contrasted organizations that ``prioritize efficiency through robots'' with workers' ``fears of having their jobs taken away''. This dimension reflects whether robots are interpreted through localized functional utility or wider socio-technical implications.

\textbf{(4) Temporal Orientation.}
This dimension reflects whether participants evaluate robots from a \textit{state-based} perspective focused on current capabilities, or a \textit{developmental} perspective that treats limitations as part of an evolving trajectory.
At the developmental end, P5 noted their perception became more positive upon realizing researchers were actively ``working on healthcare robots'', shifting their view from vague possibility to grounded feasibility; P1 emphasized that ``immature technology matures'' through continued effort. At the state-based end, concerns about current failures and malfunctions led P5 to assign a conservative score of 3 despite recognizing the technology's potential. Robots may therefore be judged primarily by present readiness or interpreted in terms of evolving potential.

\subsection{Two Spaces in a Co-evolving Loop}

To connect these interpretive dimensions to robotic systems, we synthesize design considerations from our prior co-design study \cite{bai2026towards} into a complementary robot design space.
Figure~\ref{fig:spaces} presents these four design considerations alongside the four interpretive dimensions identified above. Together, the two spaces capture how robots are designed and situated, and how they are interpreted and evaluated.

Figure~\ref{fig:spaces} specifies four bidirectional relations between
the two spaces. First, roles, tasks, and functional boundaries make
different units of evaluation visible, shaping whether people judge
specific functions or the robot as a whole; conversely, the level of
attribution guides localized functional revision or broader role
reframing. Second, embodied encounters provide the experiences and
observations people use to judge robots, while experiential evidence
informs later refinements to form and interaction. Third, deployment
conditions broaden reasoning from immediate tasks to organizational and
societal concerns, while broader concerns inform subsequent spatial,
organizational, and safety requirements. Finally, exposure to
capabilities, limitations, and development pathways frames judgments of
present readiness versus future potential, while temporal judgments
guide immediate reliability priorities or staged capability and
autonomy development. These pairings reflect recurring connections
identified across the two studies, rather than exhaustive or exclusive
one-to-one correspondences.

The two directions draw on complementary evidence: the present
follow-up study examines how robot-design considerations shape human
interpretation, while the co-design study documents how participants'
interpretations and situated experiences informed iterative design
decisions.
Viewed over time, these bidirectional relations operate within the
\textbf{co-evolving loop} illustrated in Fig.~\ref{fig:loop}. Human
needs are articulated into design requirements, which are realized
through deployed robotic systems. Situated interaction then shapes
human interpretation, which is further socially mediated and
articulated back into future design cycles.

This conceptualization builds on the task-artifact cycle in
human-computer interaction~\cite{carroll1989artifacts}, the duality
of technology \cite{orlikowski1992duality},
 and work in social robotics on the mutual shaping of robots and
society~\cite{vsabanovic2010robots}. It also aligns with recent calls
for bidirectional human-AI alignment~\cite{shen2024towards}. We extend
these perspectives to healthcare robotics by specifying and empirically
grounding the relations between robot-design considerations and human
interpretive dimensions.
This framing shifts attention from whether coexistence emerges to how
this evolving relationship should be cultivated, motivating the notion
of \textit{considerate human-robot coexistence} discussed below.

\subsection{Toward Considerate Human-Robot Coexistence}

We use \textbf{considerate} to describe a quality of coexistence in
which the design and introduction of robotic systems remain attentive
and responsive to the stakeholders in healthcare. This orientation resonates with care-centered
value-sensitive design, which emphasizes that robots for care should be
designed and evaluated in relation to the real needs and values of care
practice \cite{Wynsberghe2013DesigningRF}.
What counts as considerate is therefore specified through stakeholders'
situated needs, responsibilities, and interpretations.
Our findings further show that humans are not only \textbf{design
contributors} but also \textbf{interpreters and mediators} of how robots
are understood, introduced, and refined over time. Considerate
coexistence thus requires more than designing robots to be considerate
toward humans; it requires participatory processes through which
stakeholders articulate concerns, calibrate expectations, and support
iterative refinement as robots are integrated into care. In this sense,
considerate design extends from a property of the robot~\cite{bai2026towards}
to a property of the coexistence process itself. We derive four
implications for its realization.

\textbf{Implication 1: Make design rationale legible before deployment.}
Even when full co-design is not feasible, stakeholders should be told
what the robot is intended to do, how it fits into existing workflows,
and why key design decisions were made. Such transparency can align expectations and reduce ambiguity before stakeholders encounter the robot.

\textbf{Implication 2: Support familiarization during deployment.}
Once robots are introduced, test runs, live demonstrations, and informal
encounters can help stakeholders ask questions, observe robot behavior,
and form grounded expectations. Material, form, and role signaling can
further reduce unfamiliarity by making robots feel less intrusive and
more appropriate to the setting.

\textbf{Implication 3: Support gradual engagement and individual choice.}
Stakeholders vary in their readiness to engage with robots, especially
during early adoption. Deployment should therefore let people decide
whether, when, and how to interact, using staged exposure and opt-in
options to accommodate the different comfort levels participants raised.

\textbf{Implication 4: Emphasize functional value while respecting
professional boundaries.}
Robot integration should support care delivery rather than promote
acceptance for its own sake. Participants emphasized that healthcare
organizations remain responsible for patient care, especially in
critical conditions; robots therefore need clearly defined roles,
functional contributions, and human oversight, echoing prior
work on human-robot teamwork in healthcare~\cite{eriksen2023understanding}.

\section{Limitations and Future Work}

Because the follow-up relied on voluntary participation from a sustained
co-design cohort, the sample comprised nine in-depth interviews, which
may limit the breadth of perspectives represented. Future work should examine how interpretations evolve as systems are refined in practice, and include more diverse stakeholders
and healthcare contexts to strengthen the transferability of the perception space.

\bibliographystyle{IEEEtran}
\bibliography{main.bib}
\end{document}